\documentclass[11pt,a4paper]{article}

\usepackage{a4}
\usepackage{amsmath}
\usepackage[dvips]{graphics}
\usepackage{pstricks}
\usepackage{natbib}
\usepackage{latexsym}

\begin{document}

\title{Typical Models: Minimizing False Beliefs}
\author{Eliezer L. Lozinskii \\
School of Computer Science and Engineering \\
The Hebrew University, Jerusalem 91904, Israel \\
{\it email: lozinski@cs.huji.ac.il}}
\date{}

\maketitle

\newtheorem{definition}{Definition}[section]
\newtheorem{theorem}{Theorem}[section]
\newtheorem{example}{Example}[section]
\newtheorem{lemma}{Lemma}[section]
\newtheorem{corollary}{Corollary}[section]
\newtheorem{conjecture}{Conjecture}[section]
\newtheorem{proposition}{Proposition}[section]
\newtheorem{algorithm}{Algorithm}[section]
\newtheorem{observation}{Observation}[section]

\newcommand{\pt}{phase transition }
\newcommand{\ptt}{phase transition}
\newcommand{\pts}{phase transitions }
\newcommand{\ptss}{phase transitions}
\newcommand{\et}{\emph{et al.} }
\newcommand{\co}{$|$ \hspace{-3.5mm} $\sim$}                         

\begin{abstract}

A knowledge system $S$ describing a part of real world 
does in general not
contain complete information. Reasoning with incomplete
information is prone to errors since any belief derived from $S$ may
be false in the present state of the world. A false belief may suggest
wrong decisions and lead to harmful actions. So an important goal is
to make false beliefs as unlikely as possible. This work introduces 
the notions of \emph{typical atoms} and
\emph{typical models}, and shows that reasoning with typical models
minimizes the expected number of false beliefs over all ways
of using incomplete information. Various properties of typical models
are studied, in particular, correctness and stability of beliefs
suggested by typical models, and their connection to oblivious
reasoning.

\end{abstract}


\section{Introduction}

Let us consider a knowledge system $S$ describing a part of real
world. The knowledge contained in the system consists of data
describing properties of various objects of the world, their
mutual relationship, laws governing their behavior and evolution. For
example, consider a system $S$ of medical knowledge about the ``world''
of a hospital. $S$ contains description of diseases (their causes,
development, consequences, examination, symptoms, treatment,
prevention), information about various medicament (their composition,
therapeutic activity, dosage, directions for use, interactions, side
effects), description of the hospital (its structure,
management, location), personal data of the hospital patients (their
medical history, test results), general rules of medicine, etc. 
An important decision that has to be
made by a physician for his or her patient is determining the right
diagnosis and the best treatment. The physician may wish to consult
the vast amount of knowledge collected in the system. Will the system 
help the physician to make a right decision? This depends to a large
extent on the way the knowledge is used for deriving conclusions.

Let us define important features of $S$ and their correspondence
to the world it describes.
\vspace{-3mm}
\begin{quote}
$S$ is presented in a first order language.
%\footnote{Many existing 
%Knowledge Systems like OpenCyc \citet{cyc} are
%formulated in languages based on Predicate Calculus.}. 
$S$ is consistent having a set $MOD(S)$ of models. Each model
of $S$ is a set of ground atomic formulas expressed in terms of values
assigned to various objects and parameters of the world (such as names
of patients, quantities of medicament, etc.).

The multitude of models of $S$ reflects uncertainty regarding 
actual values of some of these parameters.
For instance, as long as neither a final diagnosis nor a
treatment of a patient $A$ is determined, there are, say, two possible
diagnoses: $D_1$ (with possible treatments $T_1$ or $T_2$) or $D_2$
(with $T_3$ or $T_4$). So $MOD(S)$ may contain four different models,
each including $D_1\; T_1$ or $D_1\; T_2$ or $D_2\; T_3$ or $D_2\; T_4$.

A set of values of all parameters of the world
(including those not presented in $S$)
determines its \emph{state}. With regard to patient $A$ the hospital
world has at least four possible states, each represented in $S$ by
one of the four models mentioned above.

Since the
available knowledge of the real world is incomplete in general, it may
happen that a model $\mu$ of $S$ represents several 
states of the world that
differ in the reality but are indistinguishable from the point of view
of the information presented in $S$. For $S$ the states represented by
$\mu$ constitute an equivalence class of possible states of the
world called a \emph{possible world} (denoted $w$ in the sequel).

$S$ describes the world \emph{faithfully} in the sense that
every possible state of the world belongs to a possible world
represented by a model of $S$, and every model of $S$ represents a
non-empty set of possible states of the world. 
From the point of view of $S$ the real world appears as a set $W$ 
of possible worlds such that there is a bijection
between $MOD(S)$ and $W$.
\end{quote}

A user of $S$ having a query whether a formula $F$ is true in the
present state of the world applies to $S$ and expects
to get an answer
based on the information stored in $S$. If $F$ (or $\neg F$) is a
logical consequence of $S$ then $S$ contains
\emph{complete} information about $F$. In this case $F$ is true
(false) in all models of $S$ and all possible worlds. Hence $F$ is
certainly true (false) in the present state.
However, if neither $F$ nor $\neg F$ follows
from $S$ then the information in $S$ regarding $F$ is
\emph{incomplete} and does not facilitate derivation of a definite
answer to the query. But this does not diminish the need or importance
of a reasonable answer. A physician cannot delay for a long time a
treatment of a patient just because he or she is not yet certain about
the final diagnosis. Travelers reaching a crossroads would not just stay
there even if they are not sure which way leads to their
destination. 

Mark Twain wrote, ``The trouble with the world is not
that people know too little, but that they know so many things that
ain't so.'' Even given an extensive knowledge of the world, its
incompleteness makes erroneous judgment inevitable.
In the present state of the world a formula 
$F$ has a certain value although this value
may be uncertain from the standpoint of $S$.
If the information about $F$ contained in $S$ is incomplete
(briefly, $F$ is incomplete in $S$), we have to find a way of reasoning
producing a \emph{belief} regarding $F$ which is \emph{credible} in
the sense that it stands a good chance of being true in the reality.
So a system of automated reasoning must
be able to answer the following query:
\begin{quote}
Given a formula $F$ and a system $S$ describing faithfully a world,
what is a most credible belief regarding the truth of $F$ in 
the present state of the world?
\end{quote}

In order to answer various multiple queries consistently a reasoner
has to choose one particular model $\mu$ of $S$
(a \emph{preferred model}) and then believe 
that $F$ is true in the reality 
iff $\mu \models F$. If $S \models F$ or $S
\models \neg F$ then the choice of $\mu$ does not matter;
however, if $F$ is incomplete in $S$ then $F$ is true in some models,
but false in the others. Which is the correct value 
of $F$ in the present state of the world? With any
choice of $\mu$ there is a non-zero probability that the belief in $F$
implied by $\mu$ is false in the present state of the world. 
So reasoning with incomplete
information is \emph{prone to errors}. 

The way of choosing the
preferred model provides a semantics for the process of 
reasoning. Whatever
this way is, errors are inevitable since the preferred model
may not fully conform to
the present state of the world. The smaller the expected number (or
the severity) of errors, the more reliable the semantics. Numerous
approaches to reasoning with incomplete information 
have been developed including
\emph{Nonmonotonic Logics} \citep{ant97, bre07, sho87} 
and methods based on \emph{Semantics of Minimal Models}
\citep{bid86, gel88, mcc80, min82, van91}. Neither of the previous work
considered minimization of the risk that beliefs sanctioned by the
proposed semantics are false in the real world. A false belief may
suggest wrong decisions and lead to harmful actions. As reasoning
errors caused by incompleteness of information are inevitable,
minimization of the number and likelihood of false beliefs becomes
practically important a goal. 

The following sections introduce the 
\emph{semantics of typical models} and show that it minimizes 
the expected number of erroneous beliefs over all ways of
reasoning with incomplete information. 

\section{Evidence}

At any moment the world is in exactly one of its possible states, so
in exactly one of possible worlds represented by the corresponding
model of $S$. Let $p(w)$ denote the probability that at a randomly
chosen moment the world is in a state belonging to a possible world 
$w \in W$
represented by a model $\mu$ of $S$. Then to every $\mu \in MOD(S)$
representing the corresponding 
$w \in W$ one may assign a probability $p(\mu) = p(w)$
such that $\sum_{\mu \in MOD(S)} p(\mu) = \sum_{w \in W} p(w) = 1$.
So the probability $p(F)$ that a formula $F$ is true in the 
present state of the world is
\begin{equation}
p(F) = \sum_{\mu \in MOD(S \cup \{F\})} p(\mu)
\end{equation}
where $MOD(S \cup \{F\}) = \{\mu | \mu \in MOD(S) \wedge \mu \models
F\}$ is the set of models of $S$ implying $F$.

If $p(F) > 0.5$ then it is reasonable to believe that $F$ is more
likely to be true than false in the present state of the world,
and the larger $p(F)$, the more credible this belief. 

The problem, however, is that in most practical cases there is no
reliable information regarding the distribution of $p(w)$. In the
absence of this information 
let us assume just for the moment that all possible worlds are
equiprobable, and sets $W$ and $MOD(S)$ are finite. Appendix B shows
a way to relax these limitations in case that certain knowledge is
available about probability of possible worlds and structure of a
given system and its domain.

The assumptions of the previous paragraph lead to the following approach.

\begin{definition}[Principle of majority of models, PMM] \label{def-PMM}
Believe that a formula $F$ is more likely to be true than false in a
state of the world 
if $F$ is true in a majority of models of $S$. The larger
the majority, the more credible the belief. $\Box$
\end{definition}

A reasonable semantics should respect the power of majority;
indeed, $F$ is true (false) in $S$ if it is true
(false) in all models of $F$.
Obeying such an unanimity, would it be
reasonable to disregard a majority of 99.9\% or even 80\%? 

As PMM suggests a belief regarding the truth value of $F$, we may say 
that the set of models of $S$ offers an \emph{evidence}
of $F$, $E(S, F)$. We would like the evidence to provide
a quantitative measure of credibility of the
corresponding belief. To normalize the value of evidence for all 
$S$ and $F$
such that $0 \le E(S, F) \le 1$, it is reasonable to require that
$E(S, F) = 1$ for $S \models F$, $E(S, F) = 0$ for 
$S \models \neg F$, and $E(S, F) + E(S, \neg F) = 1$ for all $S, F$.
More requirements are presented in \citet{loz94} leading to the
following definition.

\begin{definition} \label{def-evid}
\emph{Evidence} of $F$ in $S$:
\begin{equation}
E(S, F) = \frac{|MOD(S \cup \{F\})|}{|MOD(S)|}. \; \; \; \Box
\end{equation}
\end{definition}

PMM suggests that $F$ is true if $E(S, F) > 0.5$ or 
$E(S, F) = 0.5$ (the latter is chosen to avoid ambiguity; see also
footnote 3). Given a query regarding the truth value of $F$,
a reasoner may not only return 'true' or 'false', but also attach
the value of $E(S, F)$ to the answer to give a measure of credibility
of the latter. In cases where accepting an erroneous answer can have
very undesirable consequences, a query can require a certain level of
credibility, for example, ignoring answers with evidence less than
0.9. 

\section{Oblivious vs. non-oblivious reasoning}

In the absence of sufficient statistical information the evidence
$E(S, F)$ is regarded as an approximation of the probability 
$p(F)$ that $F$ is true in a randomly chosen possible world, 
that is the probability that the belief in the truth of $F$ is correct
in the present state of $W$. 

Consider a reasoner $R$ that forms beliefs in order to answer 
a series of queries $F_1, \ldots , F_k$. 
Denote by $R(F_i)$ his belief regarding the truth of $F_i$. 
If the reasoner computes $R(F_i)$ as his answer to $F_i$ 
without taking into account the previous beliefs 
$R(F_j)$ ($1 \le j < i$) preceding
$R(F_i)$, let us call this way of reasoning \emph{oblivious}.
Then if it turns out that there is no model of $S$ in which all 
of $R(F_1), \ldots , R(F_i)$
are true, then the beliefs of $R$ are inconsistent with $S$ which is
unacceptable. 

Oblivious reasoning with incomplete
information may lead to inconsistency. Indeed, let $M_{R(F_i)}$
denote a set of all models of $S$ 
in which $R(F_i)$ is true. Then the set of
beliefs $\{R(F_1), \ldots , R(F_k)\}$ is consistent with $S$ (and so,
holds in some state of $W$) iff 
\begin{equation}
\bigcap_{i = 1}^{k} M_{R(F_i)} \not= \emptyset.
\end{equation}
For all queries $F$ incomplete in $S$, $M_{R(F)}$ is a proper subset
of $MOD(S)$, so
the size of their intersection (3) is a monotone decreasing function 
of $k$ such that for a large $k$ condition (3) may not hold
\footnote{For instance, in Example 3.1 expression (3) holds for $k =
2$, but not for $k = 3$.}. This does
not happen if the reasoning is \emph{non-oblivious} 
such that in derivation of $R(F_i)$ all previously produced beliefs 
are taken into consideration. One way of doing so is to derive
$R(F_i)$ from $S \cup \{R(F_1), \ldots , R(F_{i-1})\}$.
In this case, however, the value of each belief depends on the
order of queries in their sequence.

\begin{example} \label{ex-nonobliv}
$S = \{a \vee b, b \vee c, c \vee a, \neg a \vee \neg b \vee 
\neg c\}$;\\
$MOD(S) = \{\{a, b, \neg c\}, \{a, \neg b, c\}, \{\neg a, b, c\}\}$.\\
Queries: $F_1 = a, F_2 = b, F_3 = c$; $k = 3$;
$E(S, a) = E(S, b) = E(S, c) = 2/3$.\\
Obliviously: $R(a) = R(b) = R(c) = 'true'$ which is inconsistent
with $S$.\\
Non-obliviously: let $S_0 = S$, $S_i = S_{i-1} \cup \{R(F_i)\}$
for $1 \le i \le k$. Then \\
$E(S_0, a) = 2/3; R(a) = 'true'; \; S_1 = \{a, b \vee c, \neg b \vee 
\neg c\}$;\\
$E(S_1, b) = 1/2; R(b) = 'true'; \;\; S_2 = \{a, b, \neg c\}$;\\
$E(S_2, c) = 0; R(c) = 'false'$. All these beliefs hold in the first model
of $S$. \;$\Box$
\end{example}

Non-oblivious reasoning requires keeping track of many previously
produced beliefs, so in general it is more time-consuming than its
oblivious counterpart. Thus it would be helpful to
determine sets of queries that can be answered obliviously in any
order without any risk of inconsistency. A trivial example is a set of
all formulas $F$ such that $S \models F$.
Subsection 4.3
presents less obvious sets allowing oblivious reasoning.

\section{Semantics of typical models}

This section introduces the basic notions of typical atoms and typical
models, and studies stability of the corresponding beliefs. 

\subsection{Typical atoms}

$S$ is supposed to be formulated in a first order language, so
using the terminology of Predicate Calculus
let the \emph{base} of $S$ be a set of all ground instances of all
atomic formulas corresponding to all predicates occurring in $S$:
\begin{equation}
Base(S) = \{P^{(k)}(t_1, \ldots , t_k)\},
\end{equation}
such \hfill that \hfill $P^{(k)}(x_1, \ldots , x_k)$ \hfill occurs
\hfill in \hfill $S$, \hfill $D_i$ \hfill is \hfill the \hfill 
\emph{domain} \hfill of \hfill $x_i$, \hfill and \hfill 
$t_i \in D_i$ \\
for $1 \le i \le k$.
\begin{definition}
For each ground atomic formula $\mathbf{a} \in Base(S)$ 
let $\mathbf{\hat{a}}$
denote the \emph{typical atom} corresponding to $\mathbf{a}$ such that
the evidence of $\mathbf{\hat{a}}$ is at least as large as that of 
$\mathbf{\neg \hat{a}}$. So\footnote{If $E(S, a) = 0.5$ then $E(S, a)
  = E(S, \neg a)$, and so any one of $a$ or $\neg a$ can be considered
a typical atom. However, in practice a proper choice 
of one of $a$ or $\neg a$ should be made based on
relevant knowledge of the real world.}
\begin{equation}
\mathbf{\hat{a}} = \left\{ \begin{array}{r l}
\mathbf{a} & if \; E(S, \mathbf{a}) \ge 0.5 \\
\neg \mathbf{a} & otherwise
\end{array} \right.
\end{equation}
For a formula $F$ we define its 
\emph{typical value} $\widehat{F}$ by substituting $F$ for $\mathbf{a}$ in 
expression (5).
\end{definition}

Evidence $E(S, a)$ is introduced in order to be used as an
approximation of $p(a)$. The larger the difference $|E(S, a) - 0.5|$,
the better this approximation. However, if $E(S, a)$ is close to
$0.5$, it may diverge from $p(a)$ even qualitatively such as 
$E(S, a) > 0.5$ but $p(a) < 0.5$. But in the absence of a
sufficient statistics regarding $p(w)$ one has to rely on $E(S, a)$
and believe that any
typical atom is not less likely to be true than false in the present 
state of the world. On the other hand, the need to avoid inconsistency 
may force a non-oblivious reasoner to adopt beliefs
in negation of some typical
atoms. So questions arising in any reasoning system intended for
answering multiple queries are:
\begin{quote}
Given a set of queries
$\{F_1, \ldots , F_k\}$, is there a state 
of the world in which beliefs
in the truth of typical values 
$\{\widehat{F}_1, \ldots , \widehat{F}_k\}$
hold for all $1 \le i \le k$, 
i.e. is there a model $m$ of $S$ such that 
$m \models \bigwedge^k_{i=1} \widehat{F}_i$? What is the value of 
evidence $E(S, \bigwedge^k_{i=1} \widehat{F}_i)$?
\end{quote}
The answer to the first question is positive if the latter evidence is
larger than zero.

Let $A^{(k)}$ be a set of $k$ literals $l$ such
that $l \in \{a, \neg a\}$, $a \in Base(S)$, and
$A^{(k)}_{\wedge} = \bigwedge_{l \in A^{(k)}} l$.
The following theorem estimates the value of evidence of 
$A^{(k)}_{\wedge}$.

\begin{theorem} \label{th-evid}
$\max(0, \alpha, \beta) \le E(S, A^{(k)}_{\wedge}) \le \min(1,
\gamma)$, where\\
\vspace{-5mm}
\begin{eqnarray}
\alpha = \sum_{l \in A^{(k)}} E(S, l) & - & k + 1,\\
\beta = 1 - \frac{2^{|Base(S)| - k}(2^k - 1)}{|MOD(S)|},
\quad \gamma & = & \frac{2^{|Base(S)| - k}}{|MOD(S)|}.
\end{eqnarray}
\end{theorem}

{\bfseries Proof}. First, we prove by induction on $k$ that
\begin{equation}
|MOD(S \cup A^{(k)})| \ge \sum_{l \in A^{(k)}} |MOD(S \cup
 \{l\})| - (k - 1)|MOD(S)|.
\end{equation}

\emph{Base}. For $k = 1$ inequality (8) holds trivially.

\emph{Step}. Let $M_1, M_2$ be subsets of $MOD(S)$, then
\begin{equation}
|M_1 \cap M_2| \ge |M_1| + |M_2| - |MOD(S)|. \nonumber
\end{equation}
If inequality (8) holds for all $1 \le i \le j$, then 
it holds for $j + 1$. Indeed, let
$A^{(j+1)} = A^{(j)} \cup \{l'\}$, then
\begin{equation}
|MOD(S \cup A^{(j+1)})| = |MOD(S \cup A^{(j)}) \cap MOD(S 
\cup \{l'\})| \ge \nonumber
\end{equation}
\begin{equation}
\sum_{l \in A^{(j)}} |MOD(S \cup \{l\})| -
(j - 1)|MOD(S)| + |MOD(S \cup \{l'\})| - |MOD(S)| = \nonumber
\end{equation}
\begin{equation}
\sum_{l \in A^{(j + 1)}} |MOD(S \cup \{l\})| - j|MOD(S)|. \nonumber
\end{equation}
Further, by (8) and since evidence is a non-negative value, 
\begin{equation}
E(S, A^{(k)}_{\wedge}) = \frac{|MOD(S \cup A^{(k)})|}{|MOD(S)|} 
\ge \max \left( 0, \sum_{l \in A^{(k)}} E(S, l) - k + 1 \right).
\end{equation}

Next, $A^{(k)}_{\wedge}$ is true in $2^{|Base(S)| - k}$
interpretations of $S$ but false in the rest of them. So
\begin{equation}
1 - \frac{2^{|Base(S)|}(1 - 2^{-k})}{|MOD(S)|} \le
E(S, A^{(k)}_{\wedge}) \le 
\min \left( 1, \frac{2^{|Base(S)| - k}}{|MOD(S)|} \right).
\end{equation}
Expressions (9) and (10) complete the proof. $\Box$

Let $\mathcal{F}^{(k)}$ be a set of $k$ formulas, and 
$\mathcal{F}^{(k)}_{\wedge} = \bigwedge_{F \in \mathcal{F}^{(k)}} F$.
If $A^{(k)}$ is replaced with $\mathcal{F}^{(k)}$ then
Theorem 4.1 implies the following 

\begin{corollary} \label{corollary}
(i) $E(S, \mathcal{F}^{k}_{\wedge}) \ge 
\sum_{F \in \mathcal{F}^{(k)}} E(S, F) - k + 1$; 

(ii) For \hfill all \hfill formulas \hfill $\phi, \psi$, \hfill if
\hfill $E(S, \hat{\phi}) > 0.5$ \hfill then \hfill $\hat{\phi} \wedge
\hat{\psi}$  \\
is consistent with $S$;

(iii) If $E(S, F) = 0.5$ call $F$ a \emph{neutral} formula. 
If there are two neutral formulas in a set $\mathcal{F}^{(k)}$
then $\mathcal{F}^{(k)}_{\wedge}$
may be inconsistent with $S$. $\Box$
\end{corollary}

\subsection{Typical models}

Let $T(S)$ denote the set of all typical atoms of $S$,
and $T(m)$ be the set of all typical atoms contained in a model $m$:
\begin{equation}
T(S) = \{\hat{a} | a \in Base(S)\}, \quad 
T(m) = \{\hat{a} | \hat{a} \in m\} = T(S) \cap m.
\end{equation}

\begin{definition} \label{def-typmod}
If there exists a model $\mu$ of $S$ such that $T(\mu) = T(S)$ then
$\mu$ is \emph{the most typical model} of $S$. For all $m \in MOD(S)$,
if there is no model $m'$ of $S$ such that $T(m) \subset T(m')$ then
$m$ is a \emph{typical model} of $S$. $\Box$
\end{definition}

A system $S$ may have no most typical model, but every $S$ has a
typical one. Indeed, every typical atom $\hat{a}$ is consistent with
$S$, so there is a model $m$ containing $\hat{a}$. 
Either $m$ is a typical model of
$S$ or there is a typical model $\mu$ such that $T(m) \subset
T(\mu)$. 

Suppose, a reasoner $R$ prefers a model $m$ assuming that it 
describes most trustfully the present possible world $w$.
Then $m$ represents the set of
$R$'s beliefs, but because of incompleteness of $S$ some of the beliefs
may be false in $w$. 
\begin{definition}
A formula $F$ is false in a possible world $w$ with probability 
$1 - p(F)$ (expression (1)). Let the \emph{erratum} $ER(A)$ of a set 
of literals $A$ 
be the expected proportion of its literals 
that are false in a randomly chosen possible world $w$. Then
taking $E(S, l)$ as an approximation of $p(l)$ we get
\begin{equation}
ER(A) = 1 - \frac{1}{|A|} \sum_{l \in A} E(S, l).
\end{equation}
\end{definition}

\begin{theorem} \label{th-typmod}
(i) For all $m \in MOD(S)$ there is a typical model $\mu$ of $S$ 
such that $ER(\mu) \le ER(m)$. 

(ii) If \hfill $\mu$ \hfill is \hfill the \hfill most \hfill typical 
\hfill model \hfill of \hfill $S$ \hfill then \hfill for \hfill all
\hfill $m \in MOD(S)$ \\
$ER(\mu) \le ER(m)$.
\end{theorem}

{\bfseries Proof.} (i) If $m$ is a typical model of $S$ then (i) holds
trivially, else there is a typical model $\mu$ such that $T(m) \subset
T(\mu)$. Denote $\delta_1 = T(\mu) - T(m) = \mu - m$, 
$\delta_2 = m - \mu$, $B = |Base(S)|$. All literals of $\delta_1$
are typical atoms.
There is a bijection between $\delta_1$ and $\delta_2$ such that to
every literal $\hat{a} \in \delta_1$ corresponds 
$\neg \hat{a} \in \delta_2$. Since for all 
$a \in Base(S)$ \hspace{1mm} $E(S, \hat{a}) \ge E(S, \neg\hat{a})$,
\begin{equation}
ER(\mu) - ER(m) = \frac{1}{B} 
\sum_{\hat{a} \in \delta_1} (E(S, \neg\hat{a}) -
E(S, \hat{a})) \le 0.
\end{equation}

(ii) If \hfill $\mu$ \hfill is \hfill the \hfill most 
typical \hfill model \hfill of \hfill $S$
\hfill then \hfill for \hfill all \hfill $m \in MOD(S)$ \\
$T(m) \subset T(\mu)$, hence $ER(\mu) \le ER(m)$. $\Box$

By Theorem 4.2, if there exists 
the most typical model of $S$ then it is the most trustworthy
one among all models of $S$.
Otherwise there is a typical model with a
minimum value of erratum among all models of $S$.

Let $ER(mtm)$, $ER(rand)$, $ER(worst)$, $E(S)$ denote respectively
the erratum of the most typical model, the expected erratum of a
randomly chosen model, the erratum of a model containing no typical
atoms, the average evidence of a typical atom of $S$. Then
\begin{equation}
E(S) = \frac{1}{B} \sum_{a \in Base(S)} E(S, \hat{a}),
\quad \quad ER(mtm) = 1 - E(S), 
\end{equation}
\begin{equation}
ER(rand) = \frac{2}{B} \sum_{a \in Base(S)} E(S, \hat{a})
(1 - E(S, \hat{a})), \quad \quad  ER(worst) = E(S)
\end{equation}
such that 
\begin{equation}
\lim_{E(S) \to 1} \frac{ER(rand)}{ER(mtm)} = 2, \quad \quad
\lim_{E(S) \to 1} \frac{ER(worst)}{ER(mtm)} = \infty.
\end{equation}

Since the most typical model of a given system would be the most
trustworthy one, it should be preferred by any rational reasoner. So
the existence of a most typical model is a practically important
characteristic of any knowledge system. 
Let $p(mtm)$ denote the probability that a given system $S$ has 
the most typical model. The probability that a randomly chosen 
model of $S$ contains all
typical atoms is $\prod_{a \in Base(S)} E(S, \hat{a})$. Then
\begin{equation}
p(mtm) = 1 - \left( 1 - \prod_{a \in Base(S)} E(S, \hat{a}) \right) ^M
\end{equation}
where \hspace{1mm} $M = |MOD(S)|$ \hspace{1mm} and \hspace{1mm}
$2^{-B} \le \prod_{a \in Base(S)} E(S, \hat{a}) \le (E(S))^B$.\\
So
\begin{equation} \label{e-pmtm}
1 - (1 - 2^{-B})^M \: \le \: p(mtm) \: \le \: 1 - (1 - (E(S))^B)^M.
\end{equation}

Expression (18) provides rather rough bounds for $p(mtm)$. Experimental
estimation of $p(mtm)$ is presented in Section 7.

\subsection{Typical kernel}

Since a system $S$ may be inconsistent with the set $T(S)$ of all its
typical atoms, it may have no most typical model. But $S$ must have a
typical model containing a subset of $T(S)$ consistent with $S$.
It would be helpful to characterize a subset of typical atoms of any
system $S$ that is necessarily consistent with $S$ regardless of
beliefs assigned to other atoms of $S$.
If for a given system this subset is
non-empty then queries about atoms of the subset can be answered
obliviously in any order.

\begin{definition} \label{def-neighb}
(i) Considering \hfill any \hfill model \hfill as \hfill a \hfill set \hfill of
\hfill literals, \hfill call \hfill two \hfill models \\
$m', m''$ $\mathbf{a}$-\emph{neighbors} if they differ only
in the value of an atom $\mathbf{a} \in Base(S)$ 
such that $\mathbf{a} \in m'$, 
$\neg \mathbf{a} \in m''$, and
$m' - \{\mathbf{a}\} = m'' - \{\neg \mathbf{a}\}$. 
Let $MN(S \cup \{\mathbf{a}\})$, $MN(S \cup \{\neg \mathbf{a}\})$ denote 
sets of all $\mathbf{a}$-neighboring models of $S$ such that 
every model of $MN(S \cup \{\mathbf{a}\})$ contains $\mathbf{a}$,
every one of $MN(S \cup \{\neg \mathbf{a}\})$ contains 
$\neg \mathbf{a}$, and to every model
of $MN(S \cup \{\mathbf{a}\})$ corresponds
exactly one $\mathbf{a}$-neighbor in
$MN(S \cup \{\neg \mathbf{a}\})$, and vice versa. 

(ii) If for a typical atom $\mathbf{\hat{a}}$ every model of $S$ 
containing $\neg \mathbf{\hat{a}}$ has an $\mathbf{a}$-neighbor
in $MN(S \cup \{\mathbf{\hat{a}}\})$, that is
\begin{equation} \label{kern-prop}
MOD(S \cup \{\neg \mathbf{\hat{a}}\}) =
MN(S \cup \{\neg \mathbf{\hat{a}}\}),
\end{equation}
then call $\mathbf{\hat{a}}$ \emph{a kernel atom}
possessing \emph{the kernel property} (\ref{kern-prop}), 
and let \emph{the typical kernel} of $S$, $tk(S)$, be the set of all
kernel atoms of $S$.
Figure 1 illustrates the kernel property. $\Box$
\end{definition}

By the kernel property, $tk(S)$ includes all atoms $b$ such that
$S \models b$ since $MOD(S \cup \{\neg b\}) = \emptyset$.

Let us say that a formula $\phi$
\emph{cancels} all models of $S$ in which $\phi$ is false.

\begin{lemma} \label{lemma}
For all kernel atoms $\hat{a}$ of $S$ and all literals
$l \not= \neg\hat{a}$, if $l$ is consistent with $S$ then $l$ is so
with $S \cup \{\hat{a}\}$.
\end{lemma}

{\bfseries Proof.} Suppose $l$ is consistent with $S$,
but inconsistent with 
$S \cup \{\hat{a}\}$, and so cancels all models of 
$MOD(S \cup \{\hat{a}\})$ including $MN(S \cup \{\hat{a}\})$.
Since $l \not= \neg \hat{a}$, $l$ cancels all models of 
$MN(S \cup \{\neg \hat{a}\})$ as well. By the kernel property of
$\hat{a}$, 
$MN(S \cup \{\neg \hat{a}\}) = MOD(S \cup \{\neg \hat{a}\})$. 
So $l$ cancels all
models of $MOD(S)$ and becomes inconsistent with $S$ --- a 
contradiction. $\Box$

\begin{figure}

\setlength{\unitlength}{1mm}

\begin{center}

\begin{picture}(110,80)

\put(15,20){\framebox(40,15){$MN(S \cup \{\neg\hat{a}\})$}}
\put(55,20){\framebox(40,15){$MN(S \cup \{\hat{a}\})$}}
\put(55,35){\line(0,1){20}}
\put(95,35){\line(0,1){20}}
\put(55,55){\line(1,0){40}}
\put(40,7.5){\line(-2,1){23}}
\put(70,7.5){\line(2,1){23}}
\put(30,45){\line(-3,-2){13}}
\put(40,45){\line(3,-2){13}}
\put(70,65){\line(-3,-2){13}}
\put(80,65){\line(3,-2){13}}
\put(47,5){$MOD(S)$}
\put(20.5,47.5){$MOD(S \cup \{\neg\hat{a}\})$}
\put(61.5,67.5){$MOD(S \cup \{\hat{a}\})$}

\end{picture}

\end{center}

\vspace{-10mm}

\caption{\label{fig-kern} The kernel property of $\hat{a}$.}

\end{figure}
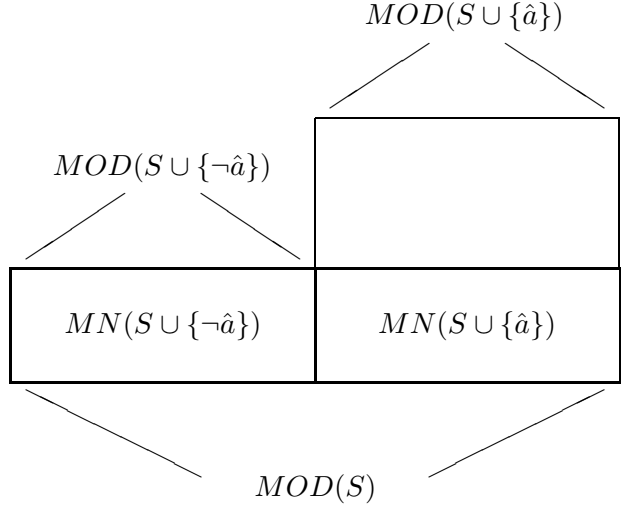

\begin{theorem} \label{th-kernel}
For all $S$, $tk(S)$ is consistent with $S$. There is no superset of
$tk(S)$ possessing this property.
\end{theorem}

{\bfseries Proof.} By induction on the serial number of 
kernel atoms of $S$ numbered arbitrarily in
$tk(S) = \{\hat{a}_1, \ldots , \hat{a}_k\}$.

\emph{Base}. Include $\hat{a}_1$ into $S$
producing $S_1 = S \cup \{\hat{a}_1\}$;
$\hat{a}_1$ is consistent with $S$,
so $S_1$ is consistent; but $\hat{a}_1$ cancels all models of $S$
containing $\neg \hat{a}_1$ such that
\begin{center}
$MOD(S_1) = MOD(S) - MOD(S \cup \{\neg \hat{a}_1\}) 
= MOD(S \cup \{\hat{a}_1\}) \not= \emptyset,$\\
$MOD(S_1 \cup \{\hat{a}_1\}) = MOD(S_1), \quad \quad
MOD(S_1 \cup \{\neg \hat{a}_1\}) = \emptyset.$
\end{center}
It turns out that for $1 < i \le k$ every $\hat{a}_i \in tk(S)$ 
is a kernel atom of $S_1$. Indeed, 

(i) by Lemma 4.1 $\hat{a}_i$
is consistent with $S_1$ since it is consistent with $S$;\\

(ii) 
\begin{center}
$MN(S_1 \cup \{\neg \hat{a}_i\}) =
MN(S \cup \{\neg \hat{a}_i\}) - MOD(S \cup \{\neg \hat{a}_1\})$;
\end{center}

(iii) since $\hat{a}_i$ is a kernel atom of $S$,
\begin{center}
$MN(S \cup \{\neg \hat{a}_i\}) = MOD(S \cup \{\neg \hat{a}_i\})$.
\end{center}

So (i) - (iii) imply \\
$MN(S_1 \cup \{\neg \hat{a}_i \}) = 
MOD(S \cup \{\neg \hat{a}_i \}) - MOD(S \cup \{\neg \hat{a}_1 \}) =
MOD(S_1 \cup \{\neg \hat{a}_i \})$. \\
Hence $\hat{a}_i$ has the kernel property in $S_1$.

\emph{Step}. Suppose kernel atoms $\hat{a}_1, \ldots , \hat{a}_i$
($1 \le i < k$)
have been included in $S$ such that 
$S_i = S \cup \{\hat{a}_1, \ldots , \hat{a}_i\}$. Then by the same
argument as above $S_i$ is consistent, and for all
$i < j \le k$ we have $\hat{a}_j \in tk(S_i)$. 
Hence $S_k = S \cup tk(S)$ is consistent.

So all kernel atoms of $S$ can be included into $S$ in any order
preserving consistency of the augmented set. However,
this may not be true regarding a non-kernel typical atom $\hat{b}$
of $S$ such that 
$\hat{b} \not\in tk(S)$. Since $b$ 
does not possess the kernel property,
$MN(S \cup \{\neg \hat{b}\}) \subset MOD(S \cup \{\neg \hat{b}\})$. 
So unlike the situation described by Lemma 4.1,
inclusion into $S$ of $tk(S)$
(or even of any literal $l$ consistent with $S$) may
cancel all models of $MOD(S \cup \{\hat{b}\})$ and of
$MN(S \cup \{\neg \hat{b}\})$. Since the latter set is just a
proper subset of $MOD(S \cup \{\neg \hat{b}\})$, we get
$MOD(S \cup tk(S)) = MOD(S \cup \{\neg \hat{b}\}) 
- MN(S \cup \{\neg \hat{b}\}) \neq \emptyset$.
Hence $\hat{b}$ is false in all models of $MOD(S \cup tk(S))$
and so inconsistent with
$S \cup tk(S)$ (or with $S \cup \{l\}$, respectively).
Thus typical kernel is the largest set of atoms 
necessarily consistent with any $S$. $\Box$

The following algorithm checks for $S$ presented in a propositional CNF 
whether a typical atom $\hat{a}$
is its kernel atom.

\begin{algorithm} \label{alg-neigh} (Clauses $c \in S$
are sets of literals; $\hat{a}$ is a typical atom of $S$).\\
1. Count $N_1 = |MOD(S \cup \{\neg\hat{a}\})|$; \\
2. Compute $S_1 = \{c - \{\neg\hat{a}\}\; |\; c \in S \; \wedge \;\hat{a} 
\not\in c\}$;\\
3. Compute $S_2 = \{c - \{\hat{a}\} \;|\; c \in S \;\wedge \; \neg\hat{a} 
\not\in c\}$;\\
4. Count $N_2 = |MOD(S_1 \cup S_2)|$; \\
5. If $N_1 = N_2$ return ``Yes, $\hat{a}$ is a kernel atom of $S$''
else return ``No''. $\Box$
\end{algorithm}

There is a bijection between $MOD(S_1)$ and $MOS(S \cup \{\hat{a}\})$,
and between $MOD(S_2)$ and $MOD(S \cup \{\neg \hat{a}\})$ 
such \hfill that \hfill
to \hfill every \hfill model \hfill $m' \in MOD(S_1)$ \\
corresponds \hfill a \hfill model \hfill 
$(m' \cup \{\hat{a}\}) \in MOD(S \cup \{\hat{a}\})$ 
\hfill and \hfill to \hfill every \hfill model \\
$m'' \in MOD(S_2)$ 
corresponds a model $(m'' \cup \{\neg \hat{a}\}) \in MOD(S
\cup \{\neg \hat{a}\})$, and vice versa. Since
$MOD(S_1 \cup S_2) = MOD(S_1) \cap MOD(S_2)$,
to every model $m \in MOD(S_1 \cup S_2)$ corresponds
$(m \cup \{\hat{a}\}) \in MN(S \cup \{\hat{a}\})$ and
$(m \cup \{\neg \hat{a}\}) \in MN(S \cup \{\neg \hat{a}\})$,
and vice versa. Hence,
$N_2 = |MN(S \cup \{\hat{a}\})| = |MN(S \cup \{\neg\hat{a}\})|$.
So line 5 of the algorithm verifies whether $\hat{a}$ possesses
the kernel property.

\begin{theorem}
For all $S$, every typical model of $S$ includes $tk(S)$.
\end{theorem}

{\bfseries Proof.} Suppose a typical model $m$ of $S$ 
does not include $tk(S)$ as it contains
a negation $\neg \hat{a}$ 
of a kernel typical atom $\hat{a} \in tk(S)$. Due to 
the kernel property of $\hat{a}$, $S$ has a model $\mu$ that 
is $\hat{a}$-neighbor of $m$ and hence contains $\hat{a}$. So 
$T(m) \subset T(\mu)$. Hence, $m$ is not a typical model --- a
contradiction. By the same argument every typical model of $S$
includes $tk(S)$. $\Box$

\subsection{Stable beliefs}

People are in constant quest for knowledge. The available knowledge
about the real world is being expanded and deepened. If new
knowledge is added to $S$, the set of models of $S$ changes, and so
the set of possible worlds $W$ changes as well. 
Indeed, the new knowledge 
changes the image of the reality portrayed by $S$ for its users. 
The corresponding changes take place in sets of 
beliefs derived from $S$ by its users.
Some beliefs regarding formulas incomplete in $S$
become more certain, but others turn out to be false.

This \hfill phenomenon \hfill makes \hfill reasoning \hfill
with \hfill incomplete \hfill information \\
\emph{nonmonotonic}:
while $S$ grows, the set of belies and
conclusions derived from $S$ may shrink. The possibility that 
some beliefs may become false is rather embarrassing and harmful. 
If a reasoner uses the
semantics of typical models, this minimizes the expected number of
beliefs that may be false in the present state of the world.
Yet the reasoner would be interested to know more: Which, if
any, of his or her beliefs are \emph{stable} in the sense that they
remain credible under some additions to the system. The set of stable
beliefs would possess a property of \emph{relative monotonicity}
with respect to these additions.

The kernel property provides the following nice quality
of stability of beliefs concerning kernel atoms.

\begin{theorem} \label{stabil}
For all $S$, all $\mathbf{\hat{a}} \in tk(S)$, 
and any formula $\phi$ that is consistent 
with \hfill $S$ \hfill and \hfill does \hfill not \hfill contain
\hfill $\mathbf{a}$ \hfill in \hfill its \hfill base,
\hfill $\mathbf{\hat{a}}$ \hfill is \hfill a \hfill 
typical \hfill kernel \hfill atom \hfill of \\
$S' = S \cup \{\phi\}$. 
So addition of $\phi$ to $S$ does not
require changing the belief 
in $\mathbf{\hat{a}}$ derived from $S$ due to 
the semantics of typical models.  
\end{theorem}

{\bfseries Proof.} \hfill Since \hfill the \hfill value \hfill 
of \hfill $\phi$ \hfill does \hfill not \hfill depend \hfill on
\hfill an \hfill assignment \hfill to \\
$\mathbf{\hat{a}} \in tk(S)$, \hfill if \hfill $\phi$ \hfill 
cancels \hfill a \hfill model \hfill $m$ \hfill of \hfill $S$
\hfill containing \hfill $\mathbf{\neg \hat{a}}$ 
\hfill then \hfill it \hfill cancels \hfill the \\
$\mathbf{a}$-neighbor of $m$
containing $\mathbf{\hat{a}}$, so  still
$MOD(S' \cup \{\mathbf{\neg \hat{a}}\}) = 
MN(S' \cup \{\mathbf{\neg \hat{a}}\}$.
Hence, $\mathbf{\hat{a}}$ retains its kernel property in $S'$. 
So beliefs in $\mathbf{\hat{a}}$ derived from $S$ and $S'$ are 
identical. $\Box$

\begin{corollary}
Let $tk(S) = \{\hat{a}_1 , \ldots, \hat{a}_k\}$, and $Base(\phi)$
denote the base of a formula $\phi$. Then $tk(S)$ is monotonic
with respect to a set of all formulas $\phi$ such that 
$Base(\phi) \cap \{a_1, \ldots, a_k\} = \emptyset$. $\Box$
\end{corollary}

\begin{example} \label{ex-stabil}
$S = \{p \vee \neg q \vee r, \:\: s \vee v, \:\: \neg q \vee r \vee
  \neg s, \:\: \neg u \vee \neg s, \:\: \neg p \vee q \vee \neg v,$ \\
$s \vee \neg v, \:\: \neg q \vee r \vee \neg u, \:\: \neg p 
\vee u \vee v, \:\: q \vee v\}$.
\end{example}

Table 1 \hfill presents \hfill data \hfill describing \hfill $S$:
\hfill $MOD(S) = \{m_1, \: m_2, \: m_3, \: m_4, \: m_5\}$; \\
$m_3 = \{\neg p, q, r, s, \neg u, v\}$ \hfill is \hfill the
\hfill most \hfill typical \hfill model \hfill of \hfill $S$ \hfill
containing \hfill its \\
typical \hfill kernel \hfill $tk(S) = \{\neg p, \, r, \, s, \, \neg u,
\, v\}$ \hfill (compare \hfill $|MOD(S \cup \{\neg \hat{a}\})|$ \hfill
with \\
$|MN(S \cup \{\neg \hat{a}\})|$
for $\hat{a} \in \{\neg p, q, r, s, \neg u, v\}$). 

To illustrate stability of kernel atoms (Theorem \ref{stabil})
let us augment $S$ with
$\phi = \{\neg p \vee \neg q \vee \neg r\}$. 
Four bottom rows of Table 1 describe
$S' = S \cup \{\phi\}$. $MOD(S') = \{m_1, \: m_2, \: m_3, \: m_4\}$
since $\phi$ cancels $m_5$.
Although $\phi$ contains kernel atom $\neg p$ and
even negation of kernel atom $r$, all
kernel atoms of $S$ remain such in $S'$:
$tk(S') = tk(S)$. So in certain cases the
stability of kernel atoms extends beyond the limits determined by 
Theorem \ref{stabil}. $\Box$

\begin{table} 
\caption{\label{tab-stab} Typical kernels of $S$ and $S'$ 
(Example \ref{ex-stabil})}
\begin{center}
\begin{tabular}{|r|c|c|c|c|c|c|}
\hline
atoms $\mathbf{a}$ of $S$ & $p$ & $q$ & $r$ & $s$ & $u$ & $v$ \\
\hline\hline
models of $S$: $m_1$ & f & f & t & t & f & t \\
$m_2$ & f & f & f & t & f & t \\
$m_3$ & f & t & t & t & f & t \\
$m_4$ & f & t & t & t & f & f \\
$m_5$ & t & t & t & t & f & t \\
\hline
typical atoms $\mathbf{\hat{a}}$ of $S$ & $\neg p$ & $q$ & $r$ & $s$ & $\neg u$
& $v$ \\
$|MOD(S \cup \{\neg \mathbf{\hat{a}}\})|$ & 1 & 2 & 1 & 0 & 0 & 1 \\
$|MN(S \cup \{\neg \mathbf{\hat{a}}\})|$ & 1 & 1 & 1 & 0 & 0 & 1 \\
$tk(S)$ & $\neg p$ & & $r$ & $s$ & $\neg u$ & $v$ \\
\hline\hline
typical atoms $\mathbf{\hat{a}}$ of $S'$ & $\neg p$ & $q$ & $r$ & $s$ & $\neg u$
& $v$ \\
$|MOD(S' \cup \{\neg \mathbf{\hat{a}}\})|$ & 0 & 2 & 1 & 0 & 0 & 1 \\
$|MN(S' \cup \{\neg \mathbf{\hat{a}}\})|$ & 0 & 1 & 1 & 0 & 0 & 1 \\
$tk(S')$ & $\neg p$ & & $r$ & $s$ & $\neg u$ & $v$ \\
\hline
\end{tabular}
\end{center}
\end{table}

\section{Typical atoms vs. intuition}

Since beliefs in the truth of typical atoms are more likely to be true
in the real world than the opposite ones, we may expect that these
beliefs should correlate with conclusions suggested by human
intuition based on life experience.
These conclusions are supposed to correlate with the
semantics of typical models better than with any other semantics
preferring models different from typical ones.
The rest of this section presents a rather simple example.

\begin{example} (A growing experience)
\begin{eqnarray}
S_0 & = & Policeman(Alex) \wedge Criminal(Bob) \nonumber \\
& \wedge & (\forall x)\{(Policeman(x) \rightarrow \neg Criminal(x)
\wedge \neg Dangerous(x)) \nonumber \\
& & \hspace{10mm} \wedge (Criminal(x) \rightarrow \neg Helpful(x))\}.
\end{eqnarray}

Suppose that life experience keeps providing additional
information $\Delta S$ characterizing policemen and criminals 
under certain conditions such that for $i > 0$
\begin{eqnarray}
\lefteqn{\Delta S_i = (\forall x)\{(Policeman(x) \wedge P\_Condition_i (x)
\longrightarrow Helpful(x))} \nonumber \\
& &  \wedge (Criminal(x) \wedge
C\_Condition_i (x) \longrightarrow Dangerous(x))\}.
\end{eqnarray}
For instance,
\begin{eqnarray*}
\lefteqn{\Delta S_1 = (\forall x)\{(Policeman(x) \wedge OnDuty(x) 
\longrightarrow Helpful(x))} \\
& & \wedge (Criminal(x) \wedge
Armed(x) \longrightarrow Dangerous(x))\}.
\end{eqnarray*}

Let us ask two questions: ``Is policeman Alex helpful?'' and
``Is criminal Bob dangerous?'' So consider queries
$F_1 = Helpful(Alex), F_2 = Dangerous(Bob)$. $\Box$
\end{example}

A common-sense intuition suggests affirmative answers to both
queries. 

Denote $S_i = S_{i-1} \wedge \Delta S_i$, and let the domain $D$ 
of all terms in $S_i$ be a finite set of names of individuals in the
community under consideration. Then from expressions (20, 21)
we get by induction on $i$
\begin{equation*}
|MOD(S_i)| = (2^i + 1)^2 (4^{i+1} + 2^{i+1} + 2)^{|D| - 2}
\end{equation*}
and
\begin{equation*}
E(S_i , Helpful(Alex)) = E(S_i , Dangerous(Bob)) = 
1 - \frac{1}{2^i + 1}.
\end{equation*}

Hence for all $i > 0$
\begin{equation*}
0.5 < E(S_i , Helpful(Alex)) = E(S_i , Dangerous(Bob)) < 1,
\end{equation*}
\begin{equation*}
\lim_{i \rightarrow \infty} E(S_i , Helpful(Alex)) = 
\lim_{i \rightarrow \infty} E(S_i , Dangerous(Bob)) = 1.
\end{equation*}

So for all $i > 0 \quad Helpful(Alex)$ and $Dangerous(Bob)$ are typical 
atoms in $S_i$ suggesting beliefs in agreement with the common-sense
intuition, and the larger $i$ the better this agreement.
By Corollary 4.1 (ii), $Helpful(Alex) \wedge Dangerous(Bob)$
is consistent with all $S_i$.
\vspace{3mm}

Noteworthy, any approach preferring a minimal model 
yields counter-intuitive beliefs in this example. Indeed, by
definition, a model $m$ is a \emph{minimal model} of $S$ if there is
no model $\mu$ of $S$ such that the set of unnegated atoms of $\mu$
is a proper subset of the set of unnegated atoms of $m$. 
For all $i \ge 0 \quad S_i$ has a single minimal model in which
all atoms except $Policeman(Alex)$ and $Criminal(Bob)$ are
negated suggesting that under all circumstances Alex is not helpful
and Bob is not dangerous --- beliefs that are hardly reasonable.

\section{Computing evidence}

Recently several algorithms have been developed for counting models
\citep{bay00, bir99, gom06, loz92, mor06, san05, thu06, wei05}
that can be employed for computing evidence. The following
algorithm (based on the algorithm CDP \citep{bir99}) has been used in this
work for computing evidence of propositional formulas.

\begin{algorithm} \label{alg-CDP}
Given $S$, let $V = \{v_1, \: v_2, \ldots , v_n\}$ be a set of all 
propositional variables of $S$.

1. Apply to $S$ the Davis-Putnam-Logemann-Loveland procedure \citep{dav62}.
Let $P^{(k)} = \{l_1, \ldots , l_k\}$ represent a sequence of
truth assignments to literals on a \emph{path} from the root of the search
tree to a node. If $P^{(k)}$ satisfies $S$, but $\{l_1, \ldots , l_{k-1}\}$
does not, call $P^{(k)}$ a \emph{satisfying path}. Let a \emph{full
  assignment} be an assignment to all variables of $S$. Any full
assignment containing a satisfying path is a model of $S$.

2. Any satisfying path $P^{(k)}$ contributes $2^{n - k}$ models to
$MOD(S)$, $2^{n - k}$ models to $MOD(S \cup \{l\})$ for every
literal $l \! \in \! P^{(k)}$, and $2^{n - k - 1}$ models
to $MOD(S \cup \{l'\})$ for every literal $l'$ such that
$l' \not\in P^{(k)}$ and $\neg l' \not\in P^{(k)}$.

3. Let $\mathcal{P}$ denote a set of all satisfying paths of $S$.
Then
\begin{equation} \nonumber
|MOD(S)| = \sum_{P^{(k)} \in \mathcal{P}} 2^{n - k}
\end{equation}
 and for all literals $l$
\begin{equation} \nonumber
|MOD(S \cup \{l\})| = 
\sum_{P^{(k)} \in \mathcal{P}\;\&\;l \in P^{(k)}} 2^{n - k} + 
\sum_{P^{(k)} \in \mathcal{P}\;\&\;l \not\in P^{(k)} \;\&\;
\neg l \not\in P^{(k)}} 2^{n - k -1}.
\end{equation}

4. For all $v \in V$, 
calculate $E(S, v) = |MOD(S \cup \{v\})| / |MOD(S)|$.
\end{algorithm}

\begin{observation}
For all literals $l$: $E(S, l) = 1$ iff $l \in P$ for all $P \in
\mathcal{P}$; $E(S, l) = 0.5$ iff $l \not\in P$ and $\neg l \not\in P$
for all $P \in \mathcal{P}$; $l$ is a typical atom if $\neg l \not\in
P$ for all $P \in \mathcal{P}$.
\end{observation}

Counting models is a hard computational task that is a \#P-complete
problem \citep{val79}. At the present state of the art of computing
counting models of $S$ 
requires a time exponential in the size
of $S$. This fact puts many knowledge collections well beyond the
computational power of the existing computers. A way to overcome this
complexity problem 
is to resort to an approximation. Appendix A presents briefly
two methods of computing a fast approximat2ion of evidence.

\section{Experiments}

Non-oblivious reasoning preserves consistency of a
set of beliefs. However, this important feature is achieved at the
expense of efficiency. Since it is necessary to take into account all
beliefs produced previously, non-oblivious reasoning is harder 
computationally than
the corresponding oblivious one. 

If a system $S$ has a most typical model then any set of
beliefs consisting of typical atoms is consistent with $S$. 
In this case beliefs 
regarding typical atoms can be produced obliviously
which makes reasoning with the most typical model efficient. 

Consider a propositional formula $S$ in CNF as a set of $C$ clauses
over $B$ propositional variables, and let $r = C / B$ denote the
clauses-to-variables ratio.

To gather information regarding existence of most typical models we
have run experiments with a program that generates random 
sets of propositional clauses and
measures their parameters relevant to this study. 

Let $p(mtm)$ be the probability that a system $S$ 
has a most typical model. The closer $p(mtm)$ to 1, the
lower the probability of inconsistency caused by oblivious 
reasoning with typical atoms of $S$.
Figure 2 displays $p(mtm)$ and $ER(mtm)$ of a set of clauses as
functions of $r$ (averaged over 10000 random sets with $B = 30, 100$).

\newpage

%\begin{figure}
%\includegraphics[4cm,8cm][14cm,18cm]{f-pmtm-r.ps}
%\vspace{-8mm}
%\begin{center}
%\caption{\label{f-pmtm-r} Probability and erratum of a most typical 
%model as a function of $r$.}
%\end{center}
%\end{figure}
%11111111111111111111111111

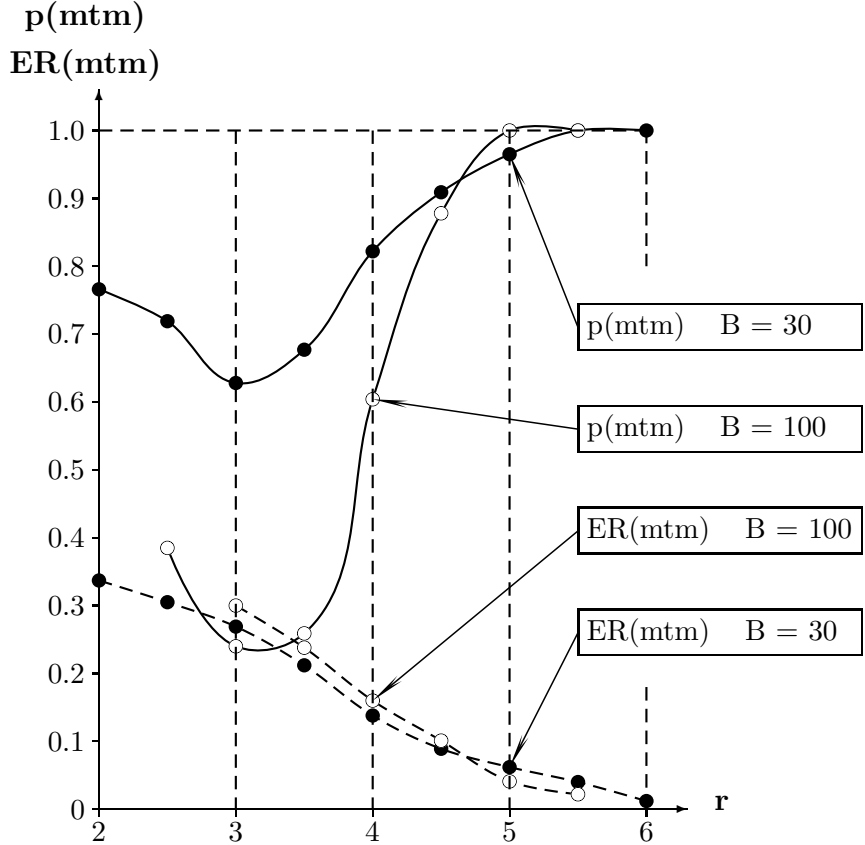
\begin{figure}
\setlength{\unitlength}{0.9mm}
\begin{center}
\begin{picture}(120,120)(-17,0)

% absciss

\put(0,0){\line(1,0){80}}
\put(80,0){\vector(1,0){6}}

\put(0,0){\line(0,-1){1}}
\put(20,0){\line(0,-1){1}}
\put(40,0){\line(0,-1){1}} 
\put(60,0){\line(0,-1){1}} 
\put(80,0){\line(0,-1){1}} 

\put(0,-2){\makebox(0,0)[t]{2}}
\put(20,-2){\makebox(0,0)[t]{3}}
\put(40,-2){\makebox(0,0)[t]{4}}
\put(60,-2){\makebox(0,0)[t]{5}}
\put(80,-2){\makebox(0,0)[t]{6}}

% ordinate
 
\put(0,0){\line(0,1){100}}
\put(0,100){\vector(0,1){6}}

\put(0,0){\line(-1,0){1}}
\put(0,10){\line(-1,0){1}}
\put(0,20){\line(-1,0){1}}
\put(0,30){\line(-1,0){1}}
\put(0,40){\line(-1,0){1}}
\put(0,50){\line(-1,0){1}}
\put(0,60){\line(-1,0){1}}
\put(0,70){\line(-1,0){1}}
\put(0,80){\line(-1,0){1}}
\put(0,90){\line(-1,0){1}}
\put(0,100){\line(-1,0){1}} 

\put(-2,0){\makebox(0,0)[r]{0}}
\put(-2,10){\makebox(0,0)[r]{0.1}}
\put(-2,20){\makebox(0,0)[r]{0.2}}
\put(-2,30){\makebox(0,0)[r]{0.3}}
\put(-2,40){\makebox(0,0)[r]{0.4}}
\put(-2,50){\makebox(0,0)[r]{0.5}}
\put(-2,60){\makebox(0,0)[r]{0.6}}
\put(-2,70){\makebox(0,0)[r]{0.7}}
\put(-2,80){\makebox(0,0)[r]{0.8}}
\put(-2,90){\makebox(0,0)[r]{0.9}}
\put(-2,100){\makebox(0,0)[r]{1.0}}

\thicklines

\put(91,2){\makebox(0,0)[t]{\bfseries{\large r}}}
\put(-2,119){\makebox(0,0)[t]{\bfseries{\large p(mtm)}}}
\put(-2,112){\makebox(0,0)[t]{\bfseries{\large ER(mtm)}}}

\put(70,70){\fbox{\begin{minipage}{35mm}
 p(mtm) \hspace{2mm} B = 30 \end{minipage}}}
\put(70,55){\fbox{\begin{minipage}{35mm}
 p(mtm) \hspace{2mm} B = 100 \end{minipage}}}
\put(70,40){\fbox{\begin{minipage}{35mm}
 ER(mtm) \hspace{2mm} B = 100 \end{minipage}}}
\put(70,25){\fbox{\begin{minipage}{35mm}
 ER(mtm) \hspace{2mm} B = 30 \end{minipage}}}

\begin{pspicture}(0,0)(130,130)

\psset{xunit=0.9mm,yunit=0.9mm}

% p(mtm) B = 30
\pscurve[showpoints=true,dotstyle=*,dotscale=1.5](0,76.6)(10,71.9)%
(20,62.8)(30,67.7)(40,82.2)(50,90.9)(60,96.5)(70,100)(80,100)

% p(mtm) B = 100
\pscurve[showpoints=true,dotstyle=o,dotscale=1.5](10,38.5)(20,24)(30,25.9)%
(40,60.4)(50,87.8)(60,100)(70,100)

% ER(mtm) B = 30
\pscurve[linestyle=dashed,showpoints=true,dotstyle=*,dotscale=1.5](0,33.7)(10,30.5)%
(20,26.9)(30,21.2)(40,13.8)(50,8.9)(60,6.2)(70,4)(80,1.2)

% ER(mtm) B = 100
\pscurve[linestyle=dashed,showpoints=true,dotstyle=o,dotscale=1.5](20,30)(30,23.8)%
(40,16)(50,10.1)(60,4.1)(70,2.2)

\psline[linewidth=0.2mm,arrowlength=5]{->}(70,71)(60,96.5)
\psline[linewidth=0.2mm,arrowlength=5]{->}(70,56)(40,60.4)
\psline[linewidth=0.2mm,arrowlength=5]{->}(70,41)(40,16)
\psline[linewidth=0.2mm,arrowlength=5]{->}(70,26)(60,6.2)

\psline[linestyle=dashed](0,100)(80,100)
\psline[linestyle=dashed](20,100)(20,0)
\psline[linestyle=dashed](40,100)(40,0)
\psline[linestyle=dashed](60,100)(60,0)
\psline[linestyle=dashed](80,100)(80,80)
\psline[linestyle=dashed](80,18)(80,0)

\end{pspicture}

\end{picture}
\end{center}

\begin{center}
\caption{\label{f-pmtm-r} Probability and erratum of a most typical 
model as a function of $r$.}
\end{center}

\end{figure}

%222222222222222222222222222
Models of any consistent set of clauses $S$ are arranged in clusters, each
determined by a satisfying path $P^{(k)}$ 
and so containing $N = 2^{B - k}$
models that have $k$ literals in common. In such a cluster the
evidence of all $k = B - \log_2 N$ common literals is $1$, and that of
each of the rest of $\log_2 N$ literals is $0.5$. Hence
the average evidence of an
atom in a cluster is $1 - (\log_2 N)/(2B)$. 
So for a system $S$ with $M$ models
$1 - (\log_2 M) / (2B)$ can be taken as an approximation of $E(S)$.
If $1 - (\log_2 M) / (2B)$ is substituted for $E(S)$ in expression (18)
then the right-hand side of (18)
has a minimum at a number of models $M_0$ determined by
equation
\begin{equation}
(1 - \phi)\ln (1 - \phi) + \frac{1}{2 \ln 2} \phi^{1 - 1/B} = 0
\end{equation}
where $\phi = (1 - (\log_2 M_0) / (2B))^B$.
Since the number of models of $S$ is a monotone decreasing function
of $r$, there is a value $r_0$ corresponding to $M_0$
at which $p(mtm)$ has a minimum as
shown in Figure \ref{f-pmtm-r}. It is worth noting that 
the erratum of a most typical model decreases with growing value of
$r$. This is in agreement with the common-sense intuition that the
more information a system contains, the more right conclusions can be
derived. 

The clauses-to-variables ratio $r$ of $S$ can be calculated
in time linear in the size of $S$, so the value
of $r$ is a convenient measure for estimating $p(mtm)$. There is
another syntactic (and so easily computable) measure of $S$ that
controls features of $S$ 
in a way similar to that of $r$. This is \emph{impurity} 
studied in \citet{loz06}.

Let \hfill $pos(v), neg(v)$ \hfill stand, \hfill respectively, \hfill 
for \hfill the \hfill number \hfill of \hfill unnegated \hfill and \\
negated occurrences of a variable $v$ in a set of clauses $S$. 
If $v$ occurs in $S$
either only unnegated or only negated ($neg(v) = 0$ or $pos(v) = 0$)
then $v$ is a \emph{pure} variable in $S$, otherwise $v$ is an
\emph{impure} one. Denote 
\begin{equation}
max(v) = \max(pos(v), neg(v)), \quad
min(v) = \min(pos(v), neg(v)).
\end{equation}
Let $imp(v) = min(v) / max(v)$ be called the \emph{impurity} of $v$, and 
$imp(S)$ stand for the impurity of $S$, that is the average impurity of
its variable:
\begin{equation}
imp(S) = \frac{1}{B} \sum_{i=1}^B min(v_i)/max(v_i) 
\end{equation}
\begin{equation}
0 \le imp(S) \le 1.
\end{equation}

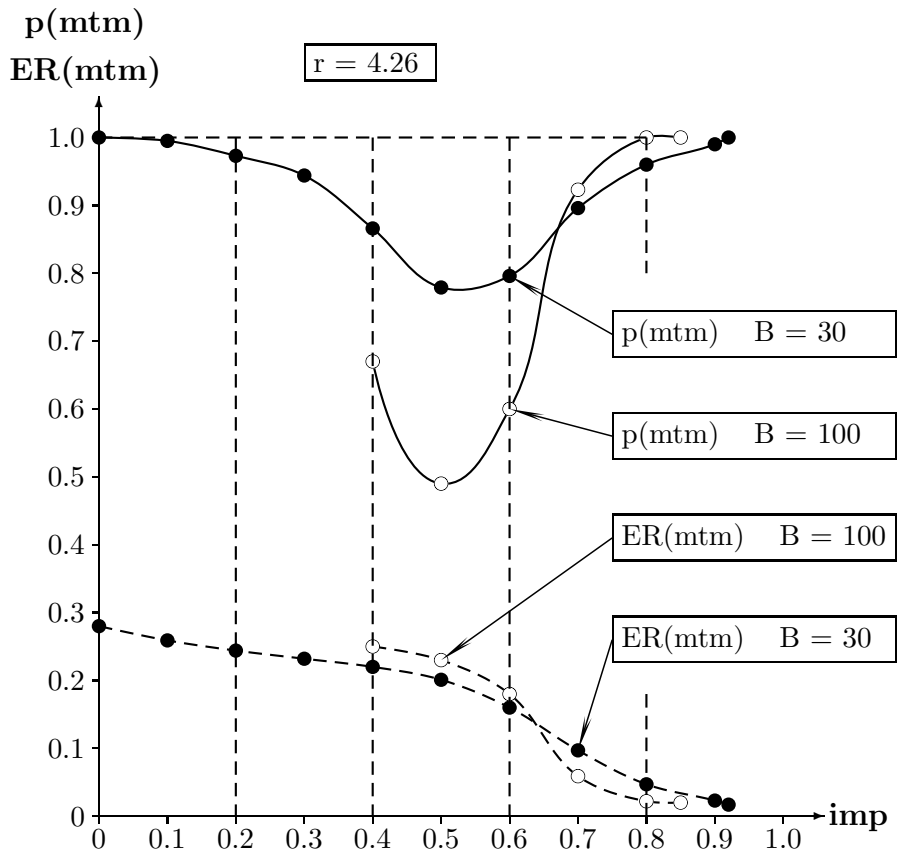
\begin{figure}
\setlength{\unitlength}{0.9mm}
\begin{center}
\begin{picture}(120,120)(-17,0)

% absciss

\put(0,0){\line(1,0){100}}
\put(100,0){\vector(1,0){6}}

\put(0,0){\line(0,-1){1}}
\put(10,0){\line(0,-1){1}}
\put(20,0){\line(0,-1){1}}
\put(30,0){\line(0,-1){1}}
\put(40,0){\line(0,-1){1}}
\put(50,0){\line(0,-1){1}}
\put(60,0){\line(0,-1){1}} 
\put(70,0){\line(0,-1){1}}
\put(80,0){\line(0,-1){1}} 
\put(90,0){\line(0,-1){1}}
\put(100,0){\line(0,-1){1}}

\put(0,-2){\makebox(0,0)[t]{0}}
\put(10,-2){\makebox(0,0)[t]{0.1}}
\put(20,-2){\makebox(0,0)[t]{0.2}}
\put(30,-2){\makebox(0,0)[t]{0.3}}
\put(40,-2){\makebox(0,0)[t]{0.4}}
\put(50,-2){\makebox(0,0)[t]{0.5}}
\put(60,-2){\makebox(0,0)[t]{0.6}}
\put(70,-2){\makebox(0,0)[t]{0.7}}
\put(80,-2){\makebox(0,0)[t]{0.8}}
\put(90,-2){\makebox(0,0)[t]{0.9}}
\put(100,-2){\makebox(0,0)[t]{1.0}}

% ordinate
 
\put(0,0){\line(0,1){100}}
\put(0,100){\vector(0,1){6}}

\put(0,0){\line(-1,0){1}}
\put(0,10){\line(-1,0){1}}
\put(0,20){\line(-1,0){1}}
\put(0,30){\line(-1,0){1}}
\put(0,40){\line(-1,0){1}}
\put(0,50){\line(-1,0){1}}
\put(0,60){\line(-1,0){1}}
\put(0,70){\line(-1,0){1}}
\put(0,80){\line(-1,0){1}}
\put(0,90){\line(-1,0){1}}
\put(0,100){\line(-1,0){1}} 

\put(-2,0){\makebox(0,0)[r]{0}}
\put(-2,10){\makebox(0,0)[r]{0.1}}
\put(-2,20){\makebox(0,0)[r]{0.2}}
\put(-2,30){\makebox(0,0)[r]{0.3}}
\put(-2,40){\makebox(0,0)[r]{0.4}}
\put(-2,50){\makebox(0,0)[r]{0.5}}
\put(-2,60){\makebox(0,0)[r]{0.6}}
\put(-2,70){\makebox(0,0)[r]{0.7}}
\put(-2,80){\makebox(0,0)[r]{0.8}}
\put(-2,90){\makebox(0,0)[r]{0.9}}
\put(-2,100){\makebox(0,0)[r]{1.0}}

\thicklines

\put(111,2){\makebox(0,0)[t]{\bfseries{\large imp}}}
\put(-2,119){\makebox(0,0)[t]{\bfseries{\large p(mtm)}}}
\put(-2,112){\makebox(0,0)[t]{\bfseries{\large ER(mtm)}}}

\put(75,70){\fbox{\begin{minipage}{35mm}
 p(mtm) \hspace{2mm} B = 30 \end{minipage}}}
\put(75,55){\fbox{\begin{minipage}{35mm}
 p(mtm) \hspace{2mm} B = 100 \end{minipage}}}
\put(75,40){\fbox{\begin{minipage}{35mm}
 ER(mtm) \hspace{2mm} B = 100 \end{minipage}}}
\put(75,25){\fbox{\begin{minipage}{35mm}
 ER(mtm) \hspace{2mm} B = 30 \end{minipage}}}

\put(30,110){\fbox{\begin{minipage}{15mm}
 r = 4.26 \end{minipage}}}

\begin{pspicture}(0,0)(130,130)

\psset{xunit=0.9mm,yunit=0.9mm}

% p(mtm) B = 30
\pscurve[showpoints=true,dotstyle=*,dotscale=1.5](0,100)(10,99.5)%
(20,97.3)(30,94.4)(40,86.6)(50,77.9)(60,79.6)(70,89.6)(80,96)%
(90,99)(92,100)

% p(mtm) B = 100
\pscurve[showpoints=true,dotstyle=o,dotscale=1.5](40,67)(50,49)%
(60,60)(70,92.3)(80,100)(85,100)

% ER(mtm) B = 30
\pscurve[linestyle=dashed,showpoints=true,dotstyle=*,dotscale=1.5](0,28)%
(10,25.9)(20,24.4)(30,23.2)(40,22)(50,20.1)(60,16)(70,9.7)(80,4.7)%
(90,2.3)(92,1.7)

% ER(mtm) B = 100
\pscurve[linestyle=dashed,showpoints=true,dotstyle=o,dotscale=1.5](40,25)%
(50,23)(60,18)(70,5.9)(80,2.2)(85,2)

\psline[linewidth=0.2mm,arrowlength=5]{->}(75,71)(60,79.6)
\psline[linewidth=0.2mm,arrowlength=5]{->}(75,56)(60,60)
\psline[linewidth=0.2mm,arrowlength=5]{->}(75,41)(50,23)
\psline[linewidth=0.2mm,arrowlength=5]{->}(75,26)(70,9.7)

\psline[linestyle=dashed](0,100)(80,100)
\psline[linestyle=dashed](20,100)(20,0)
\psline[linestyle=dashed](40,100)(40,0)
\psline[linestyle=dashed](60,100)(60,0)
\psline[linestyle=dashed](80,100)(80,80)
\psline[linestyle=dashed](80,18)(80,0)

\end{pspicture}

\end{picture}
\end{center}

\begin{center}
\caption{\label{f-pmtm-imp} Probability and erratum of a most typical model as a function of $imp$.}
\end{center}

\end{figure}

It has been shown in \citet{loz06} that while the impurity of a set of
clauses $S$ growth from 0 to 1, the probability that $S$ is
satisfiable decreases and undergoes a phase transition in the
vicinity of a certain value 
of impurity depending on $r$. The number of models of
$S$ is a monotone decreasing function of $imp(S)$ like it is as a
function of $r$. Figure \ref{f-pmtm-imp} presents $p(mtm)$ and 
$ER(mtm)$
of a set of clauses as functions of its impurity (averaged over 10000
random sets with $B = 30, 100$, $r = 4.26$, and $0 \le
imp(S) \le 0.92$). The patterns are
similar to those of Figure \ref{f-pmtm-r}.
So given $S$, both $r(S)$ and $imp(S)$ can be used for a
quick estimation of the probability that $S$ has a most typical
model. 

%\begin{figure}
%\includegraphics[4cm,8cm][14cm,18cm]{f-pmtm-imp.ps}
%\vspace{-12mm}
%\begin{center}
%\caption{\label{f-pmtm-imp} Probability and erratum of a most typical model as a function of $imp$.}
%\end{center}
%\end{figure}

\section{Conclusion}

In general, a knowledge system $S$ describing a 
real world does not
contain complete information about it. Reasoning with incomplete
information is prone to errors since any belief derived from $S$ may
turn out to be false in the present state of
the world. The smaller the expected number of false
beliefs produced by an approach to reasoning with incomplete
information, the more reliable the approach.

In regard to the main goal --- choosing a model that would represent
the reality most faithfully --- this work is close to the previous
research on reasoning with incomplete information, but presents a
completely different approach introducing typical models and showing 
that any knowledge system has a typical model that is the
most trustworthy one since it minimizes the number of false
beliefs. So if minimization of reasoning errors is important, the
semantics of typical models is the best one among all 
approaches to reasoning with incomplete information. 

\newpage

%333333333333333333333333333333

%444444444444444444444444444
\vspace{5mm}

We consider oblivious and non-oblivious reasoning. The latter unlike
the former is \emph{safe} in the sense that it does not cause
inconsistency of the set of beliefs with $S$. However, oblivious
reasoning is more efficient computationally than the corresponding
non-oblivious one.

Under the following conditions oblivious reasoning with typical
atoms is safe, and the beliefs do not depend on
the order in which they were produced:

(i) If $S$ has a most typical model then oblivious reasoning 
with all typical atoms of $S$ is safe;

(ii) Oblivious reasoning with all atoms
of the typical kernel of $S$ is safe;

(iii) The higher the probability $p(mtm)$ that $S$ has a most typical
model, the smaller the
probability that oblivious reasoning with typical atoms of $S$ is
not safe.

\section*{Acknowledgments}

Many thanks
to Amnon Barak for introducing me to the Hebrew University
Grid. The flexibility of the Grid and the power of its 600 processors
allowed performing of the experiments presented in this work.

\section*{Appendix A. Approximation of evidence}

Reasoning with typical models involves counting models. This is a
\#P-complete problem \citep{val79} presenting a highly complex
computational task that for large logic systems is beyond the power of
existing computers. One of practical ways to relax this difficulty is
using approximation.

\subsection*{A1. Credible subsets}

Given a system $S$ and a
query $F$, should it be possible to find a subset of $S$ informative
enough to provide a correct answer to $F$ with a high probability and
small enough to fit into the range of the available computing
resources, the answer to $F$ could be produced efficiently. This
approach has been studied in \citet{loz97}.

\begin{definition}
Let $L^{(1)}$ denote a subset of $S$ 
consisting of all clauses of $S$ containing a literal $L$ or
$\neg L$. Call $L^{(1)}$ the \emph{first surrounding} of $L$.
For $i > 1$ let $L^{(i)}$ denote the \emph{i-th surrounding}
of $L$, that is a set of all clauses of $S$ which either 
belong to $L^{(i-1)}$ or share a common
variable with a clause of $L^{(i-1)}$. $\Box$
\end{definition}

An $i$-th surrounding of $L$ provides an evidence $E(L^{(i)}, L)$ of
$L$ that can be considered as an approximation of $E(S, L)$ with the
\hfill \emph{approximation error} \hfill $\epsilon^{(i)}$\\
such that
$\epsilon^{(i)} = E(L^{(i)}, L) - E(S, L)$. A belief in $L$ suggested
by $E(L^{(i)}, L)$ is \emph{credible} if it is the same as that
provided by $E(S, L)$. As reported in \citet{loz97}, 
while $i$ increases, the value of $|\epsilon^{(i)}|$ decreases,
and the probability that a belief
suggested by $E(L^{(i)}, L)$ is credible approaches 1.
For most instances tested in 
\citet{loz97} the first surrounding provided credible beliefs with a
high probability, while the corresponding run time was about $10^6$
times shorter than that required for processing of the full $S$. 
The credibility of approximation increases with the second and further
surroundings along with a decrease of the run time gain.

\subsection*{A2. Comparing bounds}

Algorithm 6.1 can be used for computing upper and lower bounds of the
size of sets of models.

If a path $P^{(k)} = \{l_1, \ldots , l_k\}$ falsifies $S$ but $\{l_1,
\ldots , l_{k-1}\}$ does not, call $P^{(k)}$ a \emph{falsifying
  path}. Any full assignment containing a falsifying path is a
\emph{non-model} of $S$. Any falsifying path $P^{(k)}$ contributes
$2^{n - k}$ non-models to the set of non-models of $S$ containing a
literal $l$ for every literal $l \in P^{(k)}$, and $2^{n - k - 1}$
non-models to the set of non-models of $S$ containing $l'$ for every
literal $l'$ such that $l' \not\in P^{(k)}$ and $\neg l' \not\in
P^{(k)}$. 

Consider a run of Algorithm 6.1 starting at 
time $\tau_s$ and finishing at $\tau_f$. 
In the course of its run the algorithm discovers more and more
satisfying and falsifying paths, and accumulates models and
non-models. Let $\mathcal{M}_t (l)$, $\mathcal{N}_t (l)$
denote the number of models and non-models containing a literal $l$
counted between time $\tau_s$ and $t$.
Since $\mathcal{M}_t (l)$ and $\mathcal{N}_t (l)$ are
non-decreasing functions of $t$, this determines the following bounds
of the number of models of $S$:
\begin{equation*}
\mathcal{M}_t (l) \le |MOD(S \cup \{l\})| \le 
2^{n - 1} - \mathcal{N}_t (l);
\end{equation*}
\begin{equation*}
\mathcal{M}_t (\neg l) \le |MOD(S \cup \{\neg l\})| \le 
2^{n - 1} - \mathcal{N}_t (\neg l).
\end{equation*}

If for an atom $a$ at time $\tau(a) \le \tau_f$ 
\begin{equation}
\mathcal{M}_{\tau(a)} (a) \ge 2^{n - 1} - \mathcal{N}_{\tau(a)} (\neg a)
 \quad or \quad \mathcal{M}_{\tau(a)} ( \neg a) > 2^{n - 1} -
 \mathcal{N}_{\tau(a)} (a)
\end{equation}
then $|MOD(S \cup \{a\})| \ge |MOD(S \cup \{ \neg a\})|$, $E(S, a) \ge
0.5$, and hence the typical atom $\hat{a} = a$
or, respectively, $|MOD(S \cup \{\neg a\})| > |MOD(S \cup \{a\})|$,
$E(S, a) < 0.5$, and $\hat{a} = \neg a$. So the typical value
$\hat{a}$ can be determined already at time $\tau (a)$. At this
time the bounds give the following approximation of evidence:
\begin{equation}
\frac{\mathcal{M}_{\tau (a)} (a)}
{\mathcal{M}_{\tau (a)} (a)\!+\!2^{n-1}\!-\!\mathcal{N}_{\tau (a)}
 (\neg a)}\! \le\! E(S, a)\! \le\! 1\! -\! 
\frac{\mathcal{M}_{\tau (a)} (\neg a)}
{\mathcal{M}_{\tau (a)} (\neg a)\!+\!2^{n-1}\!-\!\mathcal{N}_{\tau
  (a)} (a)}.
\end{equation}

Let $\tau_0 (a)$ denote the earliest
time at which one of the inequalities (26) holds. It can be shown that
for all $a \in Base(S)$ if $|E(S, a) - 0.5| > 0$ then $\tau_0 (a) < \tau_f$
and 
the \hfill larger \hfill the \hfill value \hfill of \hfill 
$|E(S, a) - 0.5|$ \hfill the \hfill larger \hfill 
the \hfill run \hfill time \hfill gain \\
$(\tau_f - \tau_s)/(\tau_0 (a) - \tau_s) > 1$.
So an estimation of
evidence and determination of the corresponding typical atom can be
achieved by means of comparing bounds faster than by 
a full run of Algorithm 6.1.

\section*{Appendix B. Relaxing limitations}

So far we have assumed that all possible worlds represented by the
models of $S$ are equiprobable and the sets $W$ and $|MOD(S)|$ are
finite. This appendix shows
an example of how these limitations can be relaxed. 

\subsection*{B1. Probability of possible worlds}

In most practical cases there is no comprehensive statistical
information about 
the \hfill world \hfill sufficient \hfill for \hfill 
calculating \hfill the \hfill probability \hfill
$p(m)$ \hfill for \hfill every \hfill model \\
$m \in MOD(S)$. However, there
often is some restricted statistics regarding a subset of objects and
events of the world. For instance, 
suppose the prior probabilities of certain possible worlds are
known (as all the possible worlds are mutually exclusive, their mutual
conditional probabilities are 0). Let $\vec{M}$ be the set of models
of $S$ representing possible worlds with known probability, and denote
$p(\vec{M}) = \sum_{m \in \vec{M}} p(m)$. Then assuming that all 
possible worlds
with unknown probabilities are equiprobable, we get
\begin{equation}
E(S, F) = (1 - p(\vec{M})) \frac{|MOD(S \cup \{F\}) - \vec{M}|}
{|MOD(S) -  \vec{M}|} + \sum_{m \in (MOD(S \cup \{F\}) \, \cap \,
\vec{M})} p(m).
\end{equation}

If no prior probabilities of possible worlds are known such that
$\vec{M} = \emptyset$, then expression (28) 
becomes identical to that of
Definition 2.2. In another special case, if prior probabilities are
given for all possible worlds such that $\vec{M} = MOD(S)$ and
$p(\vec{M}) = 1$, then the
evidence $E(S, F)$ amounts to the probability $p(F)$.

\subsection*{B2. Infinite sets of models}

More research has to be done to extend the notion of evidence to
systems with infinite sets of models. Here is one possible approach.

Since the set of predicate symbols occurring in a first-order system
$S$ is finite, the reason for infiniteness of the set of its models is
the infiniteness of the domain of its terms\footnote{In particular,
  Herbrand domain of $S$ becomes infinite if $S$ contains function
  symbols or existential quantifiers producing Skolem functions.}. 
Let $D$ be an infinite
enumerable domain of $S$, $d$ denote a finite subset of $D$, and 
$S^{(d)}$ stand for the original system $S$ for which the original
domain $D$ is replaced with $d$.
Then $S$ can be viewed as a limit of $S^{(d)}$ while $d$
approaches $D$. The set of models $MOD(S^{(d)})$ is finite allowing the
following definition. 
\begin{definition}
Given $S$ and its domain $D$,
let $d_1, d_2, \ldots$ be a sequence of finite subsets of
$D$ such that $\lim_{i \to \infty} d_i = D$. Then
the evidence of a formula $F$ in $S$ is
\begin{equation}
E(S, F) = \lim_{i \to \infty} E(S^{(d_i)}, F) = \lim_{i \to \infty}
\frac{|MOD(S^{(d_i)} \cup \{F\})|}{|MOD(S^{(d_i)})|}
\end{equation}
if the latter limit exists. $\Box$
\end{definition}

Applicability of this definition depends on the nature of $S$, $D$ and
$F$, and on a proper construction of the sequence of finite subsets of
$D$ for computing the limit of $E(S^{(d_i)}, F)$.

\begin{example}
$S = (\forall x)\{(P(x) \to R(x)) \wedge (Q(x) \to R(a))\}$,
and the domain of $x$ is the set of all natural numbers.
\end{example}

Let us define $d_i = \{1, \ldots , a+i\}$. Then in $S^{(d_i)}$
we have:

If $R(a)$ is false then $P(a)$ is false and for all
$x \in d_i \quad Q(x)$ is false; for every value of $x \in (d_i - \{a\})$
the clause $P(x) \to R(x)$ has 3 satisfying assignments; so 
$|MOD(S^{(d_i)} \cup \{\neg R(a)\})| = 3^{a+i-1}$.

If $R(a)$ is true then 2 assignments satisfy $P(a) \to R(a)$ and
$Q(x) \to R(a)$ for all $x \in d_i$, and 3 assignments satisfy 
$P(x) \to R(x)$ for all $x \in (d_i - \{a\})$; so 
$|MOD(S^{(d_i)} \cup \{R(a)\}) = 2^{a+i+1}3^{a+i-1}$.

Hence,
\begin{equation}
|MOD(S^{(d_i)})| = (2^{a+i+1} + 1)3^{a+i-1}, \quad 
E(S^{(d_i)}, R(a)) = (2^{a+i+1})/(2^{a+i+1} + 1).
\end{equation}

A similar calculation gives \hspace{5mm}
$E(S^{(d_i)}, P(a)) = 2^{a+i} \, / \, (2^{a+i+1} + 1)$; \\
for all $x \in (d_i - \{a\})$ \hspace{5mm}
$E(S^{(d_i)}, P(x)) = \frac{1}{3}, \quad E(S^{(d_i)}, R(x)) =
\frac{2}{3}$;\\
for all $x \in d_i$ \hspace{5mm}
$E(S^{(d_i)}, Q(x)) = 2^{a+i} \, / \, (2^{a+i+1} + 1)$.\\

In the limit \hspace{1mm} $i \to \infty$ \hspace{1mm} we 
get \hspace{5mm}
$E(S, P(a)) = \frac{1}{2}, \quad E(S, R(a)) = 1$; \\
for all natural $x \not= a$ \hspace{5mm}
$E(S, P(x)) = \frac{1}{3}, \quad E(S, R(x)) = \frac{2}{3}$; \\
for all natural $x$ \hspace{5mm}
$E(S, Q(x)) = \frac{1}{2}. \quad \Box$

\end{document}